\documentclass{article}

\usepackage{microtype}
\usepackage{graphicx}
\usepackage{subfigure}
\usepackage{booktabs} 
\usepackage{xcolor}         
\usepackage{amsfonts}       
\usepackage{nicefrac}       
\usepackage{amsmath}
\usepackage{float}
\usepackage{graphicx}
\usepackage{subfigure}
\usepackage{algorithm}
\usepackage{algorithmic}

\usepackage{hyperref}

\newcommand{\methodName}{SONAR}

\DeclareMathOperator*{\argmax}{arg\,max}
\DeclareMathOperator*{\argmin}{arg\,min}

\title{\methodName{}: Joint Architecture and System Optimization Search}
\usepackage[accepted]{mlsys2022}

\mlsystitlerunning{\methodName{}: Joint Architecture and System Optimization Search}

\begin{document}

\twocolumn[
\mlsystitle{\methodName{}: Joint Architecture and System Optimization Search}

\mlsyssetsymbol{equal}{*}

\begin{mlsysauthorlist}
\mlsysauthor{Elias Jääsaari}{cmu}
\mlsysauthor{Michelle Ma}{cmu}
\mlsysauthor{Ameet Talwalkar}{cmu}
\mlsysauthor{Tianqi Chen}{cmu,octo}
\end{mlsysauthorlist}

\mlsysaffiliation{cmu}{Carnegie Mellon University}
\mlsysaffiliation{octo}{OctoML}

\mlsyscorrespondingauthor{Elias Jääsaari}{ejaeaesa@andrew.cmu.edu}

\vskip 0.3in

\begin{abstract}
There is a growing need to deploy machine learning for different tasks on a wide array of new hardware platforms. Such deployment scenarios require tackling multiple challenges, including identifying a model architecture that can achieve a suitable predictive accuracy (architecture search), and finding an efficient implementation of the model to satisfy underlying hardware-specific systems constraints such as latency (system optimization search). Existing works treat architecture search and system optimization search as separate problems and solve them sequentially. In this paper, we instead propose to solve these problems jointly, and introduce a simple but effective baseline method called \methodName{} that interleaves these two search problems. \methodName{} aims to efficiently optimize for predictive accuracy and inference latency by applying early stopping to both search processes. Our experiments on multiple different hardware back-ends show that \methodName{} identifies nearly optimal architectures $30\times$ faster than a brute force approach.

\end{abstract}
]

\printAffiliationsAndNotice{}

\section{Introduction}
\label{introduction}

There is an increasing demand for deploying machine learning to a diverse set of constrained hardware devices, including edge devices or new hardware platforms, which are poorly supported by existing deep learning frameworks or optimized operator libraries. Such deployment scenarios involve multiple stages. In the \textit{model development} stage, we aim to identify a highly accurate model for a given task. In the context of deep learning, this involves iteratively training and tuning individual models, and also searching for a suitable model architecture itself, a process called \textit{neural architecture search} (NAS)~\cite{pham2018efficient, liu2018darts, li2020random}.

Next, in the \textit{model deployment} stage, we aim to deploy the resulting model to production. A key component of this stage is \textit{system optimization search}, where we search for a target-specific implementation of the model, optimizing the performance of a model with respect to system metrics such as inference latency or memory usage, the former of which is the focus of our paper. This search is an iterative process over possible tensor programs' loop order, vectorization, and parallelization patterns to implement the model~\cite{chen2018learning, baghdadi2021deep,Halide_AutoSched}.

\begin{figure*}[!t]
  \centering
  \includegraphics[width=\textwidth]{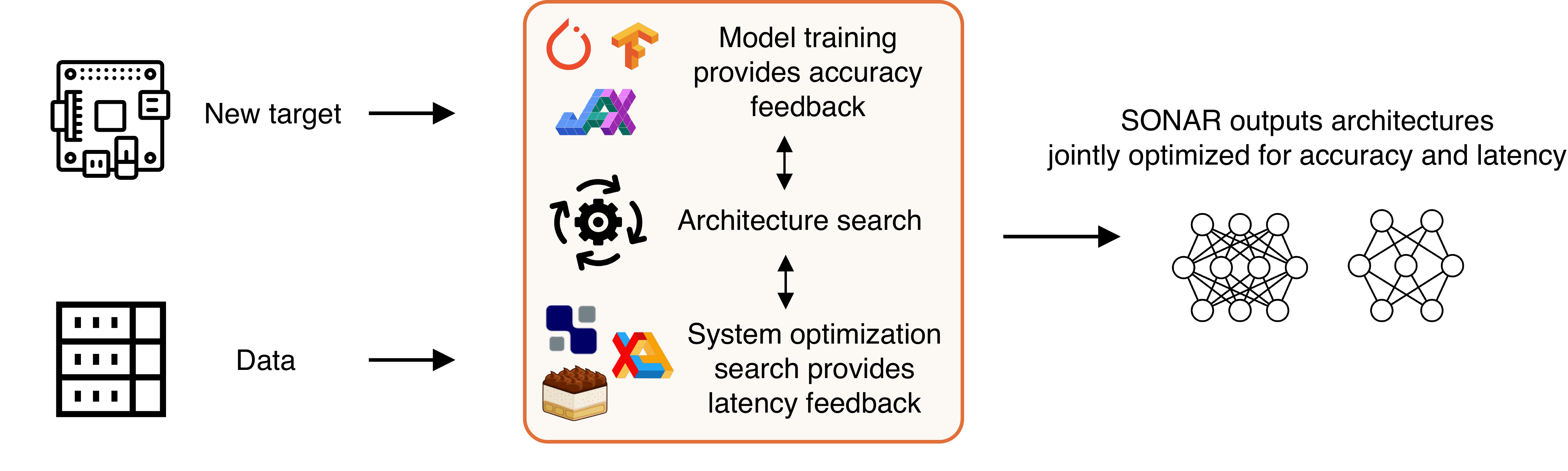}
  \caption{In SONAR, system optimization search is interleaved within the architecture search process, providing latency feedback that matches optimized latency on deployment. SONAR therefore outputs models that are jointly optimized for accuracy and latency.}
  \label{fig:overview}
\end{figure*}

Existing work treats NAS and system optimization search as sequential procedures. This sequential search strategy first executes the model development stage on a new platform using a proxy metric such as the number of FLOPs (often the only available metric) to rank models by latency, and the resulting models are subsequently optimized in the model deployment stage. The system optimization search stage is performed after the model development stage, costing additional time. Moreover, the latency proxy might not match the actual optimized latency on deployment, potentially resulting in suboptimal models.

In this work, we instead aim to jointly optimize accuracy and inference latency by interleaving NAS and system optimization search. By performing system optimization search within the architecture search process, we can execute both searches simultaneously. Since in practice the model deployment targets a device (e.g. an edge device) that is distinct from the hardware the training is performed on, system optimization search and model training can be performed in parallel, potentially allowing joint optimization at little to no extra cost added to the model development stage. While a brute force approach is simple to implement by fully evaluating each objective, it is also highly inefficient.

We propose \textbf{\methodName{}} ({\bf S}ystem {\bf O}ptimization and {\bf N}eural {\bf Ar}chitecture Search) as a simple yet effective baseline method for joint optimization of accuracy and inference latency. Our key insight is that we can perform this joint optimization efficiently by applying early stopping to both problems in parallel. Early stopping has been shown to be successful in neural architecture search and hyperparameter tuning by dynamically allocating the number of training epochs to different models~\cite{li2017hyperband}. We observe that early stopping is also applicable to system optimization search by dynamically allocating the number of tensor programs used to evaluate model latency.

\methodName{} repeatedly, over a series of rounds, allocates  resources in parallel for optimizing both the accuracy and the latency of a set of network candidates, and based on these results chooses promising candidates to allocate further resources to. This simple and general method allows SONAR to have the following desirable properties:
\begin{itemize}
\item {\bf Accuracy and efficiency}: \methodName{} is able to obtain direct feedback on the hardware-specific system metrics. By using early stopping to speed up the evaluation of both objectives, our experiments show that \methodName{} identifies nearly optimal architectures 30 times faster than brute force.
\item {\bf Simplicity}: \methodName{} is simple to implement, and does not require e.g. building a cost model or any additional adjustments for different models.
\end{itemize}

In addition, \methodName{} is independent of the model family, system optimization framework, and training techniques, making it also applicable for use with further advances in these areas. Furthermore, \methodName{} can be applied to e.g. tuning the hyperparameters of a network in addition to its architecture. We present variants of \methodName{} for two different objectives (Pareto frontier and latency threshold). \methodName{} can also be extended with different search techniques for e.g. filtering or proposing new candidate architectures.

Our experiments on image classification tasks using multiple different hardware platforms (Raspberry Pi 4 B, a MIPS-based camera platform, Apple M1 Pro GPU) show that \methodName{} can speed up the search by a factor of 30 compared to brute force while still identifying nearly optimal architectures. To achieve this improvement, SONAR does not require collecting any performance data prior to the search.

The remainder of the paper is organized as follows. In Section 2, we discuss the relation of our work with previous approaches to system optimization and hardware-aware neural architecture search. Section 3 formalizes our problem for the two possible end goals described earlier, and presents variants of \methodName{} for both scenarios. Finally, in Section 4 we present an empirical evaluation of our method on multiple hardware platforms.

\begin{figure*}[!ht]
  \centering
  \includegraphics[width=0.95\textwidth]{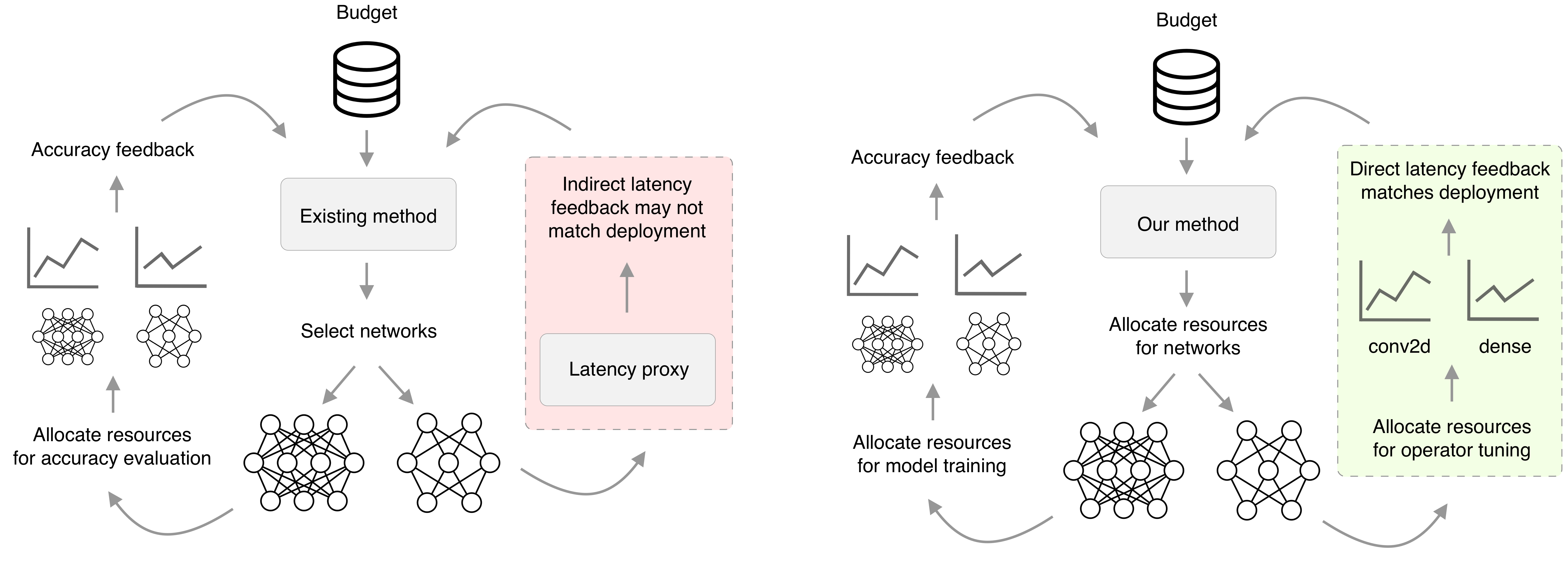}
  \caption{SONAR works by allocating resources for different networks, and then for accuracy and latency evaluation within those networks. Compared to sequential methods that use a latency proxy, resulting in inaccurate latency feedback compared to actual deployment, our method selectively allocates resources for finding jointly optimal architectures.}
  \label{fig:algo_overview}
\end{figure*}

\section{Related work}

\paragraph{Neural architecture search}{
  The single-objective neural architecture search problem for finding the most accurate architecture for a given task has been studied extensively in recent years; we refer to~\citet{elsken2019neural} for a comprehensive overview. Subsequently, several NAS methods have been proposed for constrained and multi-objective settings~\cite{elsken2018efficient, lu2021neural,lu2020nsganetv2}. Such methods can be applied for constrained inference latency by considering platform-independent metrics such as the number of floating point operations (FLOPs) as one of the constraints.
  
  As different hardware platforms can favor different architectures~\cite{li2021hw}, several methods that target a desired inference latency on a specific hardware platform have been proposed. Early work on hardware-aware neural architecture search (HW-NAS) measured the inference latency of proposed architectures directly on the target hardware~\cite{tan2019mnasnet}. Subsequent work has sped up the search by using latency predictors~\cite{wu2019fbnet, cai2019once}. In comparison to our approach, these methods aim to predict the unoptimized latency of a model using an existing deep learning framework, as opposed to the optimized latency after performing system optimization, which may result in suboptimal architectures. Furthermore, such deep learning frameworks or optimized operator libraries are not available for many platforms common in for example edge devices. For a comprehensive survey on HW-NAS, we refer to~\citet{benmeziane2021comprehensive}.
  
  Model compression techniques such as quantization~\cite{gholami2021survey,wang2019haq, wang2020apq} and network pruning~\cite{blalock2020state} can be used to compress models for efficient inference; these methods can be used to augment the
  architecture search space in \methodName{}.
  
  \citet{hernandez2016designing} propose to jointly design the neural architecture and the neural accelerator; more advanced methods have been developed subsequently~\cite{abdelfattah2020best, shi2020learned}. \citet{lin2021naas} also consider a particular limited set of compiler mappings in addition to the hardware. While in this paper we do not consider also designing the hardware, our method can also be adapted to this setting.
  
}

\paragraph{Automatic system optimizations for machine learning}{
  Several techniques have been proposed for optimizing the system performance of a single model. A \textit{machine learning compiler} takes as input a model as a computational graph and uses a search procedure, such as local search or a genetic algorithm, to find an efficient tensor programs for the model. These tensor programs implement tensor operators such as convolution and matrix multiplication. Such recent machine learning compilers include Halide~\cite{Halide_AutoSched}, TVM~\cite{chen2018learning}, XLA~\cite{kaufman2020learned}, and Tiramisu~\cite{baghdadi2021deep}. The program search space in these methods can be vast including optimizations such as vectorization and parallelization, and searching for the best tensor program can take time similar to that taken by training the model. Our method can use
  these frameworks as iterative system optimization search subcomponents.
}

\paragraph{Early stopping}{
  Early stopping can be used to speed up the accuracy evaluation of different models by training for example only on a small subset of the training data or for fewer epochs. Such an approach was proposed for hyperparameter tuning by~\cite{jamieson2016non} with accuracy as the single objective; this was subsequently improved by the Hyperband~\cite{li2017hyperband} algorithm. Hyperband has been adapted for the multi-objective setting~\cite{schmucker2020multi}; our method works similarly but takes into account the parallel and nested structure of the objectives. We also present a variation of the algorithm for the latency threshold setting. Furthermore, in this paper we make the case that early stopping can also be used for evaluating the inference latency by considering the latency after a limited number of system optimization search iterations.
}

\section{\methodName}

\subsection{Problem formalization}

\begin{figure}[!b]
  \centering
  \includegraphics[width=0.5\textwidth]{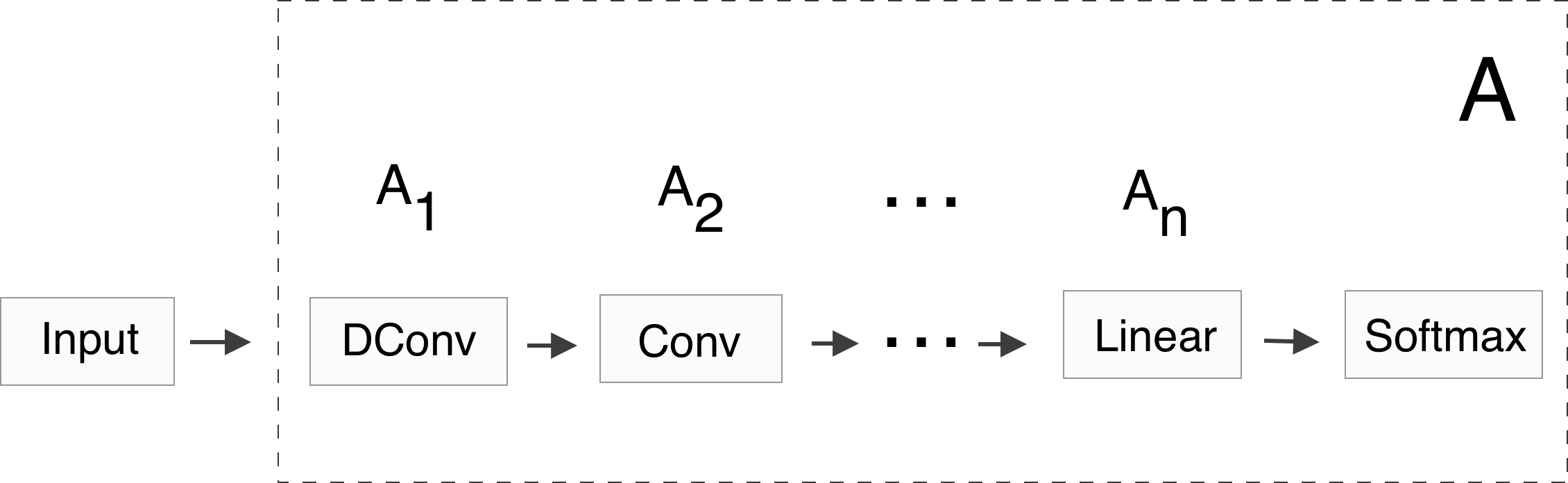}
  \caption{In the common approach taken by deep learning compilers, optimizing the inference latency of a neural architecture $A$ can be decomposed into optimizing the subgraph architectures $A_1, \dots, A_n$ independently. The latency of $A$ can then be estimated as the sum of the latencies of the subgraphs---in our experiments we find that this assumptions works well in practice.}
  \label{fig:computational_graph}
\end{figure}

Our problem has two objectives: the accuracy and the latency. We make the assumption that the latency of an architecture $A$ can be decomposed into the latencies of its $n$ subgraphs $A_1, \dots, A_n$, and thus searching for the best program for the whole architecture can be decomposed into finding the best tensor program for each subgraph (see Figure~\ref{fig:computational_graph}). This is also the approach implemented in system optimization search frameworks~\cite{chen2018learning}.

Our objective vector can now be written as
\begin{align*}
  \begin{array}{ll} & \begin{bmatrix}
  \text{Acc}_\text{valid}(w, A) & \text{Lat}(A)
\end{bmatrix}^\top, \text{ where} \\[1.5ex]
  \text{ } & w \in \underset{u \in \mathbb{R}^d}{\argmax}\, \text{Acc}_\text{train}(u, A) \\[1.8ex]
              & \text{Lat}(A) = \text{Lat}(p_1, A_1) + \cdots + \text{Lat}(p_n, A_n) \\[1.8ex]
                & p_m \in  \underset{q \in \mathcal{P}}{\argmin}\, \text{Lat}(q, A_m) \text{ for all } m = 1, \dots, n \\[1.8ex]
              \end{array}
\end{align*}

Evaluating the validation accuracy $\text{Acc}_\text{valid}(w, A)$ of an architecture $A$ involves an optimization problem over the weights $w$ of the architecture on the training set, and valuating the latency $\text{Lat}(A)$ of an architecture $A$ involves an optimization problem over the tensor programs $p_1, \dots, p_n$ implementing the subgraphs $A_1, \dots, A_n$ of $A$ (Figure~\ref{fig:computational_graph}). We evaluate the latency of each tensor program directly on the target device.

  To obtain the total latency of an architecture, we sum the latencies of the subgraphs. This assumption is only used to derive the algorithm; in practice we can measure the total latency on the target device if the assumption does not hold. In our experiments we use the TVM~\cite{chen2018tvm} compiler that considers e.g. operator fusion and in this case we find that the assumption holds well (see Section~\ref{subsec:proxy}).

Since our objective is two-dimensional, there is not a single solution that maximizes the above problem. Instead, we consider two possible formalizations that are applicable for different use cases.

\paragraph{Pareto formalization}{
In the Pareto version, we are interested in a set of Pareto-optimal solutions satisfying different accuracy--latency trade-offs, called the \textit{Pareto frontier}. An architecture is on the Pareto frontier if there does not exist another architecture with both higher accuracy and lower latency. The quality of a Pareto frontier is commonly measured using the hypervolume indicator~\cite{emmerich2005emo}. Our goal is to find a set of architectures as close to the true Pareto frontier as possible. 
}

\paragraph{Latency threshold formalization}{
In this formalization, we consider a deployment scenario that has a hard constraint on the maximum inference latency $\nu$.
We want to find the most accurate architecture not exceeding this threshold:

\begin{align*}
  \begin{array}{ll} &
    \underset{A \in \mathcal{A}}{\argmax}\, \text{Acc}_\text{valid}(w, A), w \in \underset{u \in \mathbb{R}^d}{\argmax}\, \text{Acc}_\text{train}(u, A) \\[2.5ex]
\text{s.t.} & \text{Lat}(A) \leq \nu, \text{ where} \\[1.8ex]
              & \text{Lat}(A) = \text{Lat}(p_1, A_1) + \cdots + \text{Lat}(p_n, A_n) \\[1.8ex]
                & p_m \in  \underset{q \in \mathcal{P}}{\argmin}\, \text{Lat}(q, A_m) \text{ for all } m = 1, \dots, n \\[1.8ex]
              \end{array}
\end{align*}
This formulation also encourages an algorithm to evaluate more architectures specifically near the latency threshold.
At a given threshold, this formulation might result in a more accurate architecture than using the Pareto version with the same budget. 
}

\subsection{\methodName{}}

We now present two variants of a simple yet effective baseline, \methodName{}, to solve the problem in the two settings described previously. We take an approach similar to the successive halving algorithm~\cite{jamieson2016non} and apply it to our multi-objective problem.
During the execution, \methodName{} maintains a set of promising architectures which are allocated further resources for evaluation over a series of rounds.

We emphasize that \methodName{} is a simple baseline. \methodName{} can be combined with other approaches to speed up the algorithm, such as using an accuracy predictor or a latency predictor~\cite{dudziak2020brp} to pre-filter network candidates, or using Bayesian optimization to choose new network candidates~\cite{falkner2017combining}.

To simplify the presentation, we use $S$ to denote an initial candidate set that contains all the possible architectures in the search space. In the case of a large architecture search space, we can pick the initial candidate set uniformly at random from the search space following previous approaches~\cite{li2020random} and gradually expand our 
search space in an outer loop as in~\citet{li2017hyperband}.

\begin{figure*}[!ht]
  \centering
  \includegraphics[width=.99\textwidth]{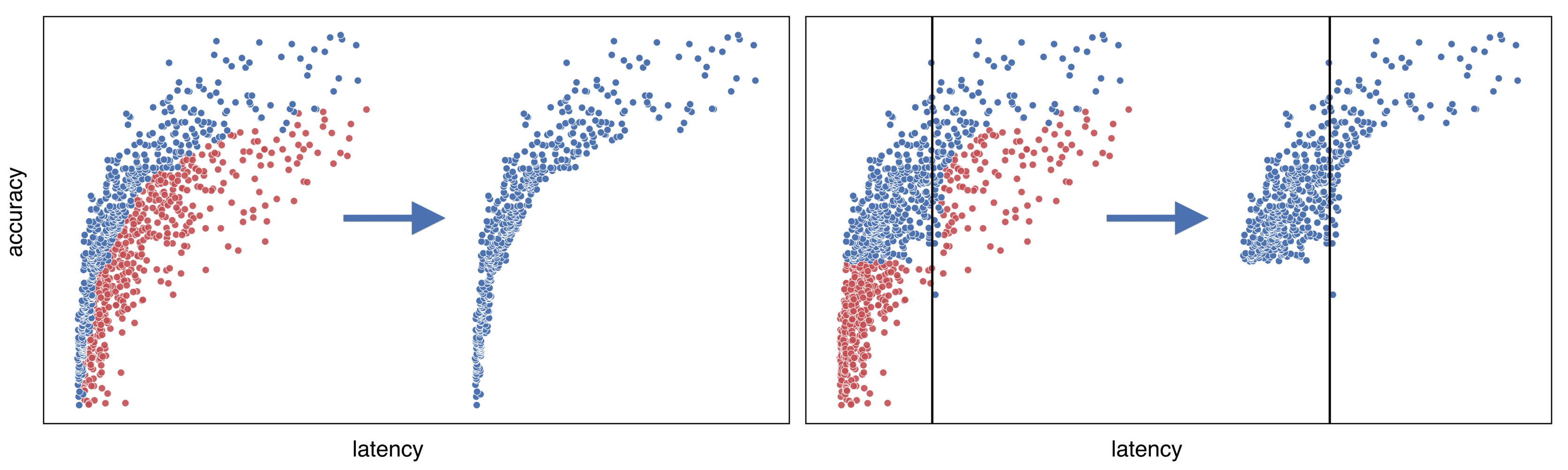}
  \caption{To eliminate suboptimal architectures from consideration, the Pareto version (left) of our algorithm repeatedly chooses the Pareto frontier of the current architectures until at most half of the architectures have been chosen. The latency threshold version (right) uses the same algorithm for architectures that exceed the latency threshold (vertical line), and simply eliminates the half of the architectures with the worst accuracy among the architectures below the threshold. Potentially more accurate architectures that do not yet meet the latency threshold have the possibility of moving below the latency threshold with further system optimization.}
  \label{fig:selection_overview}
\end{figure*}

\begin{algorithm}[tb]
   \caption{\methodName{}}
\begin{algorithmic}
   \STATE {\bfseries Input:} Budget $B$, candidate architectures $S$
   \STATE Initialize $S_0 = S$
   \FOR{$k = 0$ {\bfseries to} $\lceil \log_2(n) \rceil - 1$}
   \STATE $r_k = \lfloor \frac{B}{|S_k|\lceil \log_2(n) \rceil} \rfloor$
   \STATE Extract set of unique subgraphs $A_k$ from $S_k$
   \STATE In parallel;
   \STATE \quad \textsc{train\_networks}$(S_k, r_k)$
   \STATE \quad \textsc{optimize\_subgraphs}$(A_k, r_k)$
   \STATE $S_{k + 1} = \emptyset$
   \WHILE{$|S_{k + 1}| \leq \frac{1}{2}|S_{k}|$}
   \STATE Let $F$ be the Pareto frontier of $S_k \setminus S_{k + 1}$
   \STATE $S_{k + 1} = S_{k + 1} \cup F$
   \ENDWHILE
   \ENDFOR
   \STATE {\bfseries return} $S_{\lceil \log_2(n) \rceil}$
\end{algorithmic}
\label{alg:pareto}
\end{algorithm}

Figure~\ref{fig:algo_overview} gives an overview of our method.
To narrow down the initial candidate set to the final result, our algorithm uses partial results to speed up the evaluation of both the accuracy and inference latency objectives. For a single objective, partial evaluation can be used to obtain an estimate of the value of the objective by allocating a limited budget (number of iterations in  optimizations) to evaluate each model. The successive halving algorithm adapts this idea and, repeatedly over a series of rounds, eliminates the worse performing half of the architectures from consideration, while the remaining models are evaluated with a doubled budget.

We use partial evaluation to obtain estimates for both the accuracy and the inference latency of all the models in the current candidate set. For accuracy, we allocate resources to train the models for a further number of epochs, and use the maximum validation accuracy so far as the estimation. For latency, we allocate resources to add more auto-tuning rounds of each subgraphs and use the sum of the best latencies as the estimation.

In practice, we can obtain estimates of the objectives by allocating resources for them in parallel since (1) the objectives are independent for a given architecture, as is the case with accuracy and the kind of the system metrics that we target, and (2) that the target platform on which the system optimizations are evaluated on is a separate device (e.g.\ an edge device), and not the hardware accelerator that the training is performed on. Our method accounts for this fact. We can also adapt the algorithm to prioritize among the objectives under a sequential setting.

After obtaining estimates for our objectives, we eliminate the worse half of the architectures from consideration. Since our problem is multi-objective, we define the worse half differently for our two settings, Pareto and latency threshold, and present a separate algorithm for both. 

\paragraph{Pareto setting}{
To extend the successive halving algorithm for multiple objectives, we use non-dominated sorting (NDS)~\cite{srinivas1994muiltiobjective}. In NDS, each point in a multi-dimensional set of points $\mathcal{D}$ is assigned to a set $F_k$ depending on its rank $k$. We define $F_k$ to be the set of Pareto-optimal points when $k - 1$ Pareto frontiers have been repeatedly removed from $\mathcal{D}$. That is, we set
\begin{align*}
    F_1 &= \text{Pareto\_front}(\mathcal{D}) \\
    F_2 &= \text{Pareto\_front}(\mathcal{D} \setminus F_1) \\
    F_3 &= \text{Pareto\_front}(\mathcal{D} \setminus (F_1 \cup F_2)) \\
    \vdots
\end{align*}
To eliminate the worse performing half of models from consideration, we repeatedly remove the architectures with the highest rank until at least half of the architectures have been eliminated.
}

We describe the Pareto version of \methodName{} in Algorithm~\ref{alg:pareto}. The main loop implements the modified successive halving procedure (see Figure~\ref{fig:selection_overview} for an illustration) that repeatedly removes badly performing models from consideration. Within the loop, resources are allocated in parallel for evaluation of both the accuracy and the latency of the current candidate set of architectures. \textsc{The Train\_networks} function trains each architecture in $S_k$ for a further number of epochs as allowed by the resources $r_k$, while the \textsc{Optimize\_subgraphs} function (detailed below) optimizes the unique subgraphs in $A_k$ for a further number of candidate tensor programs as allowed by $r_k$.

\paragraph{Latency threshold setting}{
If we are only targeting a particular latency threshold, the Pareto version might miss a more accurate architecture that is not on the Pareto frontier, but is more accurate than any of the models not exceeding the latency threshold. We also do not need to keep evaluating architectures whose latencies are below the threshold but have low accuracy. However, the algorithm cannot simply discard all architectures whose latency is above the threshold on first evaluation, since further system optimizations can move the architecture below the latency threshold.

The latency threshold version of \methodName{} is similar to the Pareto version, differing only in how the worse half of the architectures are eliminated. After the evaluation is complete on each round, we divide the architectures into two sets: the set $L$ consists of architectures that are below the latency threshold after the round, and the set $R$ consists of architectures that are above it. For the architectures in $L$, we can ignore the latency objective and simply eliminate the half of the architectures with the smallest accuracies. For $R$, i.e.\ the architectures that are above the threshold but might potentially move below it in the future, we use the same Pareto frontier elimination algorithm as before; see Figure~\ref{fig:selection_overview} for an illustration. 

}

\paragraph{Optimizing subtasks in system optimization}

 Since different subgraphs can contribute to the total latency at differing magnitudes, and the optimization processes of these tasks can converge at different rates, uniform allocation of resources for tuning the different subgraphs can be suboptimal. Instead, we optimize a given subgraph by considering a batch of $\beta$ tensor programs for the subgraph at a time. Our algorithm is independent of the system optimization search process; for example AutoTVM~\cite{chen2018learning} searches over possible program candidates with different loop ordering, tiling, vectorization etc.
 
 We prioritize the allocation of resources to subgraphs whose total latency decreased the most during the last batch. This allows subgraphs that contribute the most to the latency to be optimized more, while also stopping optimization for subgraphs whose latencies are no longer improving.
 
 The algorithm maintains a priority queue of all the subgraphs with their improvement in latency during the last measurement batch as the priority. This is so that (1) we optimize subgraphs whose latencies contribute the most to the total latency more and (2) subgraphs whose latencies are no longer improving are not optimized further. We assume that when dequeueing a task from the priority queue, ties are broken uniformly at random. This ensures that if none of the tasks improved in their latency during the last iteration, further tasks are considered for possible improvement at random. The algorithm depends on the parameter $\beta$---in our experiments we set $\beta=64$.
 
\section{Evaluation}

This section aims to answer the following questions:

\begin{itemize}
    \item How well does \methodName{} match brute force search? (\S~\ref{subsec:sonar})
    \item How does \methodName{} compare to inference proxy methods? (\S~\ref{subsec:proxy})
    \item How does \methodName{} behave under different compute time budgets? (\S~\ref{subsec:budgets})
\end{itemize}

\subsection{Experimental setup}

\paragraph{Models and data}{
In all experiments, we use a MobileNetV3 like search space~\cite{howard2019searching} since we target primarily edge devices where low inference latency is a key factor. Table~\ref{table:search_space} summarizes our search space. We search over different input resolutions, width multipliers, expansion factors, kernel sizes and the number of layers in each stage of the network, resulting in a search space of 1024 models. As mentioned previously, we can also consider much larger search spaces by picking an initial candidate set e.g. uniformly at random. We use the CIFAR-10 data set in all experiments.
}
\begin{table}[t]
\caption{Architectural parameters of the MobileNetV3 like search space used in the experiments.}
\label{table:search_space}
\vskip 0.15in
\begin{center}
\begin{small}
\begin{sc}
\begin{tabular}{ll}
  \toprule \\
  Parameter & Possible values \\ 
 \midrule \\
  Resolution & \{128, 160, 192, 224\} \\
  Width multiplier & \{0.25, 0.50, 0.75, 1.00\} \\
  Expansion ratio & \{3, 6\} \\
  Depth of each stage & \{2, 3\} \\
  \bottomrule
\end{tabular}
\end{sc}
\end{small}
\end{center}
\vskip -0.1in
\end{table}

\begin{figure}[!ht]
  \centering
  \includegraphics[width=0.48\textwidth]{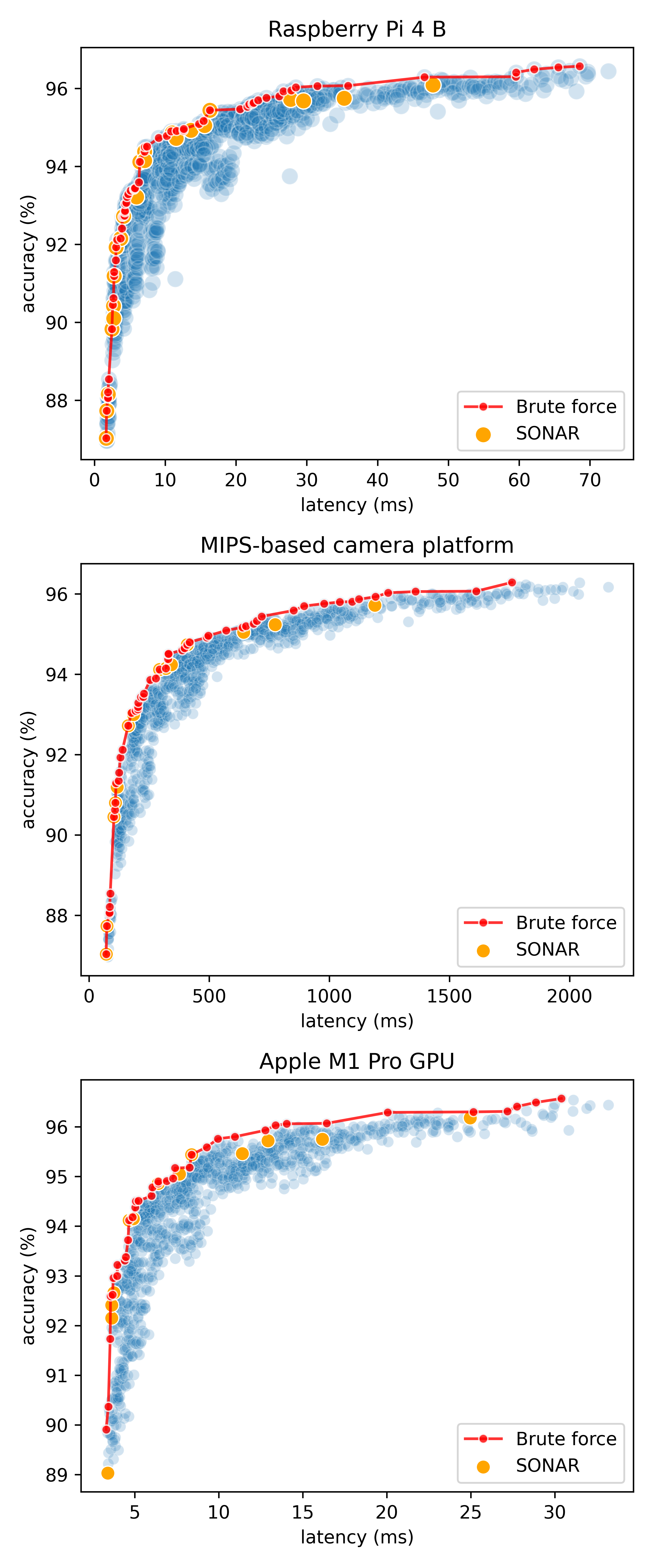}
  \caption{\methodName{} (with a time budget of two days) nearly matches a brute force search in Pareto optimality on different hardware targets (Apple M1 Pro GPU, Raspberry Pi 4, MIPS-based camera platform) in terms of Pareto optimality. However, compared to brute force search, SONAR is 34$\times$ faster.}
  \label{fig:platforms}
\end{figure}

\paragraph{Experimental environment}{
  As target hardware platforms, we use a Raspberry Pi 4 Model B (Cortex-A72 CPU), a MIPS-based camera platform (T20 CPU), and an Apple M1 Pro GPU. For the Apple M1 Pro, we always use batch size 32 for inference to account for the small size of the models. For model training, we use PyTorch 1.9. For system optimization search, we leverage the auto-scheduler framework~\cite{zheng2020ansor} in the TVM deep learning compiler~\cite{chen2018tvm} in all of the experiments for iterative system optimization of the different tasks in a network. The tasks correspond to different subgraphs of a network. Note that our method is agnostic to the choice of the deep learning compiler and can also use other frameworks for iterative system optimization search.
}

\begin{table}[t]
\caption{Rank correlation of different latency metrics versus final optimized latency using Kendall's $\tau$ coefficient on different hardware targets. (* $=$ not available for the platform)}
\label{table:kendall_tau}
\vskip 0.15in
\begin{center}
\begin{small}
\begin{sc}
\begin{tabular}{l l l l}
  \toprule
  Metric & RPi & Camera & M1 \\ 
 \midrule
  FLOPs          & 0.90 & 0.92 & 0.89 \\
  PyTorch        & 0.55 & *  & 0.90 \\
  Tensorflow     & 0.94 & *  & 0.85 \\
  SONAR (2 days) & 0.96 & 0.97 & 0.96 \\
  SONAR (4 days) & 0.97 & 0.97 & 0.97 \\
  \bottomrule
\end{tabular}
\end{sc}
\end{small}
\end{center}
\vskip -0.1in
\end{table}

\subsection{SONAR evaluation}\label{subsec:sonar}

We first evaluate the difference between the possible optimal networks found using brute force search and SONAR. For the brute force method, we evaluate the 1024 models in our search space on the CIFAR-10 data set by both fully training each model and fully tuning each model's inference latency on each of the target plaforms. We evaluate the inference latency (after optimization) of each model directly on the target platform. The total time to perform brute force is 68 days when computed based on time spent on a machine with an AMD Ryzen Threadripper 3970X CPU and two NVIDIA RTX 3070 GPUs available for model training.

In Figure~\ref{fig:platforms}, we compare these results to \methodName{} with a time budget of two days on each of the target devices. \methodName{} finds models that are nearly as optimal as those found by a brute force search, but can do so by using only two days of compute as opposed to the 68 days required by brute force.

\subsection{Latency proxy evaluation}\label{subsec:proxy}

In this section, we evaluate the viability of an even simpler baseline method than SONAR. Such a baseline method utilizes a latency proxy: compute the Pareto frontier based on the latency proxy information, and optimize only the resulting Pareto frontier. To evaluate this method, we use Kendall's $\tau$ coefficient to measure the rank correlation of different proxy methods with the optimized latency. The rank correlation can range from $-1$ to $1$, with 1 meaning identical correlation.

As proxy methods, we use the number of FLOPs, the inference latency with PyTorch, as well as the inference latency with TensorFlow (TensorFlow Lite on the Raspberry Pi and the camera). We note that the latter DL frameworks are not optimized or even available for many hardware back-ends, especially on edge devices and new platforms---for example, these frameworks are not available for the MIPS platform used by the camera, and bringing PyTorch support for the M1 GPU took approximately 18 months of developer effort. This often leaves FLOPs as the only possible proxy method on edge devices or new platforms.

\begin{figure}[!t]
  \centering
  \includegraphics[width=0.5\textwidth,clip=yes,trim=20 0 0 0]{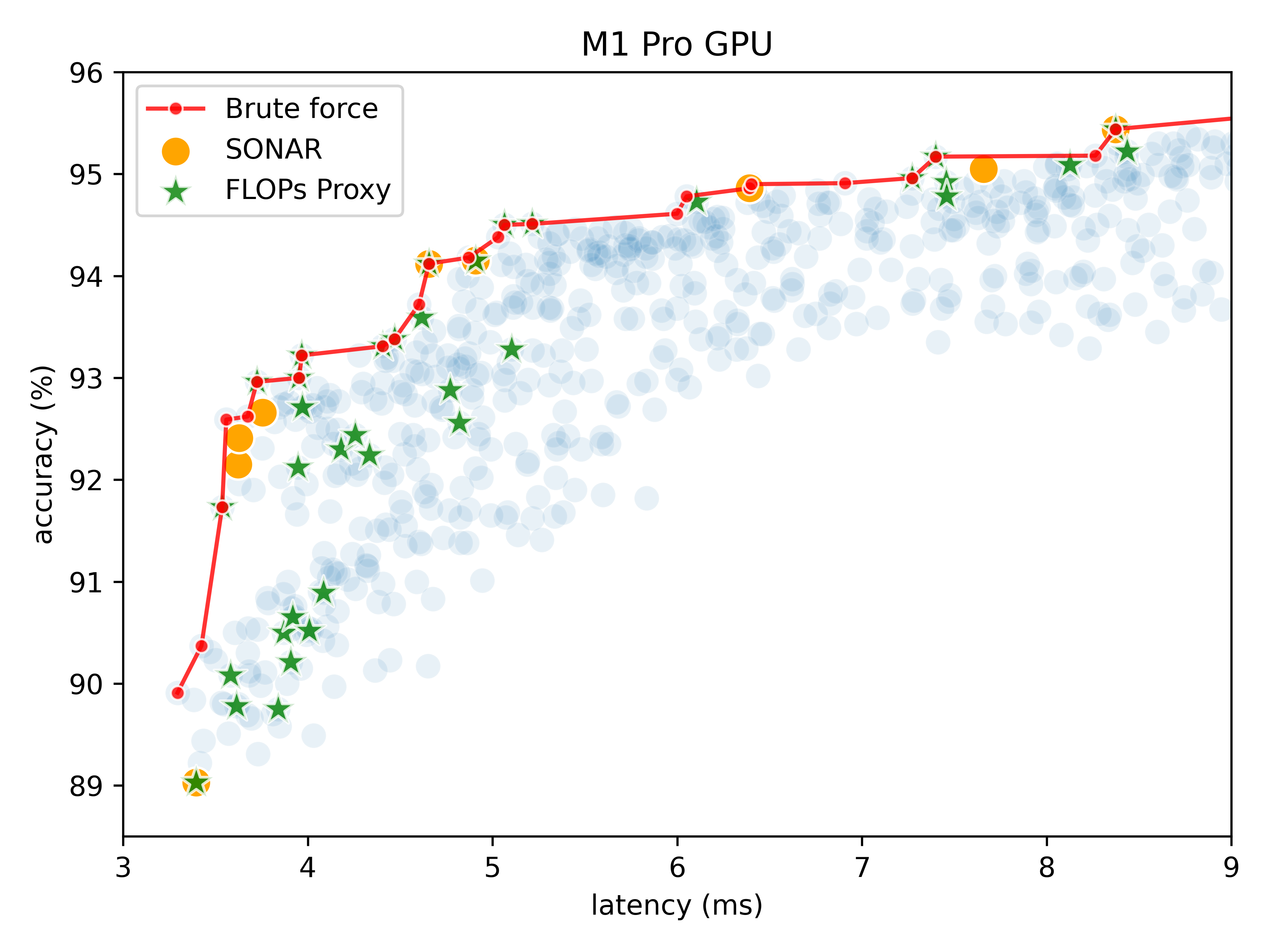}
  \caption{Architectures considered Pareto optimal by brute force, SONAR, and using FLOPs as a proxy metric on the M1 Pro GPU. In some cases, utilizing a latency proxy such as FLOPs (in practice the only available latency proxy on a new hardware back-end) can result in the search process finding many suboptimal architectures that also need to be subsequently optimized for latency.}
  \label{fig:m1_gap}
\end{figure}

The results are shown in Table~\ref{table:kendall_tau}. We also include the rank correlation of the subgraph latency sum estimation used by SONAR after 2 and 4 days (for a budget of 4 days). While SONAR achieves the best rank correlation, the proxy methods also achieve generally high rank correlation. However, they can still result in suboptimal models as shown in Figure~\ref{fig:m1_gap}. It is also in some cases difficult to know \textit{a priori} which latency proxies do not work well, e.g.\ the PyTorch inference latency results in poor rank correlation on the Raspberry Pi.

\subsection{Time budget evaluation}\label{subsec:budgets}
\begin{figure}[!t]
  \centering
  \includegraphics[width=0.5\textwidth]{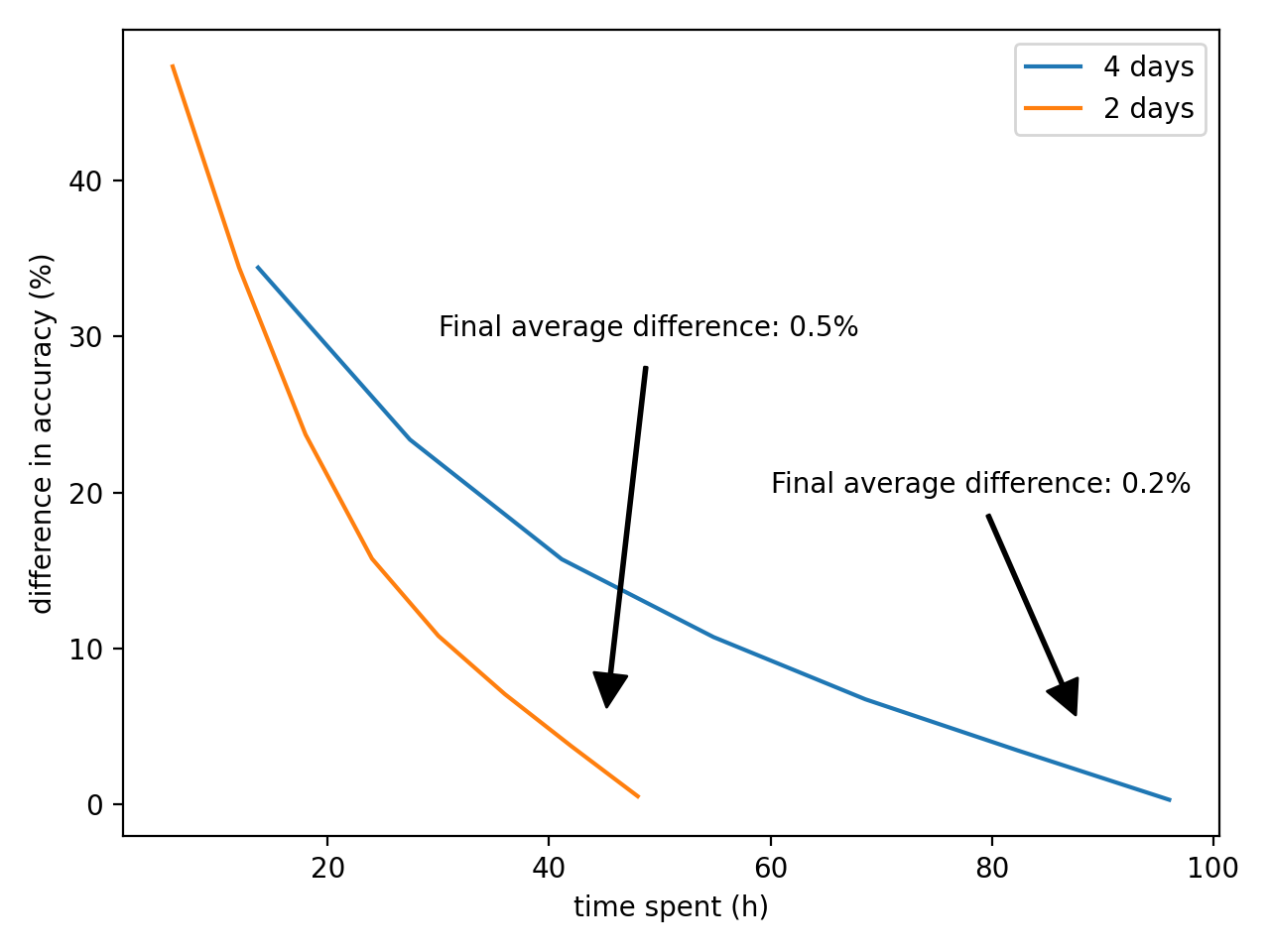}
  \caption{The progress of \methodName{} in the Pareto setting on the Raspberry Pi. At each time step, for each true Pareto-optimal network, we take the difference in its accuracy and the best accuracy for the networks found by \methodName{} so far that do not exceed the latency of the Pareto-optimal network. We then plot the average of these differences across time spent by the algorithm.  The final average difference in accuracy for the budget of 2 days is 0.5\%, while the final average difference for the budget of 4 days is 0.2\%.}
  \label{fig:progress}
\end{figure}

In this section, we evaluate whether doubling the available compute budget for \methodName{} allows it to find additional models on the Pareto frontier. We quantify these results during the execution of \methodName{} using the Raspberry Pi 4 B. At different time steps, for each true Pareto-optimal network, we take the difference of its accuracy and the best accuracy among networks that have been evaluated by our method at that time step and do not exceed the latency of the Pareto-optimal network. We then average these differences in accuracy. Figure~\ref{fig:progress} shows the progress as a function of time for two time budgets, 2 days and 4 days. The final average difference in accuracy for the budget of 2 days is 0.5\%, while the final average difference for the budget of 4 days is 0.2\%. In contrast, a brute force search takes 68 days, meaning \methodName{} is capable of finding models $34\times$ faster on average within 0.5\% of the optimal accuracy, and $17\times$ faster on average within 0.2\% of the optimal accuracy.

\section{Conclusion}

This work considers the joint optimization of accuracy and latency of neural architectures, and proposes \methodName{} as an efficient baseline method for the problem. \methodName{} applies the early stopping paradigm to both evaluating the accuracy and the inference latency of candidate models, providing direct feedback that matches the deployment on both objectives. This approach allows SONAR to be used even on platforms not supported by existing deep learning frameworks. Our experimental results demonstrate that \methodName{} can be $30\times$ faster than a brute force search approach.

While in this paper we consider the case where the objectives are accuracy and latency, we  note that \methodName{} can also be applied to other problems in a similar setting---it only requires multi-fidelity information, that is, the status of the search process after a certain amount of resources have been spent on the optimization process.

We further note that our method is independent of the system optimization search process, and can be integrated with new advancements in this area, potentially resulting in even bigger advantages. We also hypothesize that applying our approach can yield even bigger differences when considering models from different model families simultaneously, or considering model families and target platforms that allow for more optimizations.  

\section*{Acknowledgements}

This work was supported in part by the National Science Foundation grants IIS1705121, IIS1838017, IIS2046613, IIS2112471, the Real Time Machine Learning (RTML) DARPA project, and funding from Meta, Morgan Stanley and Amazon. Any opinions, findings and conclusions or recommendations expressed in this material are those of the author(s) and do not necessarily reflect the views of any of these funding agencies.

\bibliography{refs}
\bibliographystyle{mlsys2022}

\end{document}